\pdfoutput=1
\documentclass[11pt]{article}

\usepackage[]{acl}

\usepackage{times}
\usepackage{latexsym}
\usepackage{multirow} 
\usepackage{booktabs} 
\usepackage{tabularx} % For flexible table width

\usepackage[T1]{fontenc}
\usepackage[utf8]{inputenc}

\usepackage{microtype}

\usepackage{inconsolata}

\usepackage{graphicx}

\title{Quantifying the Risks of Tool-assisted Rephrasing to Linguistic Diversity}

\author{Mengying Wang \\
  University of Konstanz\\
  \texttt{mengying.w.wang@gmail.com} \\\And
  Andreas Spitz \\
  University of Konstanz\\
  \texttt{andreas.spitz@uni.kn} \\}

\begin{document}
\maketitle
\begin{abstract}
Writing assistants and large language models see widespread use in the creation of text content. While their effectiveness for individual users has been evaluated in the literature, little is known about their proclivity to change language or reduce its richness when adopted by a large user base. In this paper, we take a first step towards quantifying this risk by measuring the semantic and vocabulary change enacted by the use of rephrasing tools on a multi-domain corpus of human-generated text.
\end{abstract}

%%%%%%%%%%%%%%%%%%%%%%%%%%%%%%%%%%%%%%%%%%%%%%%%%%%%%%%%
%%%%%%%%%%%%%%%%%% Introduction %%%%%%%%%%%%%%%%%%%%
%%%%%%%%%%%%%%%%%%%%%%%%%%%%%%%%%%%%%%%%%%%%%%%%%%%%%%%%

\section{Introduction}

Writing assistants such as Grammarly or Quillbot are widely used in writing tasks by native and non-native speakers alike. To aid the user in the composition of text, writing assistants (WATs) provide sophisticated and comprehensive functions, such as grammar correction, spell-checking, and specialized rephrasing and embellishment. More recently, large language models (LLMs) are increasingly being integrated into writing assistants~\cite{fok2023can}, including Grammarly and Google's Smart Compose~\cite{DBLP:conf/kdd/ChenLBCZLTWDCSW19}. These advanced WATs offer substantial assistance in the writing process~\cite{roe2023review}, but also introduce challenges such as hallucinations or inconsistent content and style~\cite{ariyaratne2023comparison,kacena2024use,gero2022sparks}. After the introduction of ChatGPT, language models can also be -- and are -- used as writing assistants directly due to their chat functionality and ease of use.

Despite their widespread use, however, there is a notable lack of systematic quantitative investigations into how the use of writing assistants alters the produced text. Existing studies tend to focus on the effectiveness and accuracy of WATs~\cite{gayed2022exploring}, but discount the downstream impact that a pervasive adoption of such systems might have on the style and diversity of the language we use in written day-to-day communication. 

\textbf{Contributions.}
In this paper, we address this research gap with a quantitative analysis of the effect that the use of (semi)automated rephrasing tools has on the produced text. We experiment with four traditional WATs and four LLMs to determine the effect of tool-assisted rephrasing from the perspective of text diversity. Using token-level and vector-level metrics, we provide a comprehensive overview of the ways in which these tools modify text -- and where worries about linguistic diversity may or may not be warranted.

%%%%%%%%%%%%%%%%%%%%%%%%%%%%%%%%%%%%%%%%%%%%%%%%%%%%%%%%
%%%%%%%%%%%%%%%%%% Related Work %%%%%%%%%%%%%%%%%%%%
%%%%%%%%%%%%%%%%%%%%%%%%%%%%%%%%%%%%%%%%%%%%%%%%%%%%%%%%

\section{Related Work}

Academic work on writing assistants is relatively scarce, and focuses predominantly on measuring the effectiveness of WATs as a tool for improving writing efficiency, the accuracy of grammar, or for spell checking~\cite{gayed2022exploring}, and we are unaware of any studies of language diversity as a result of writing assistant usage. Furthermore, analyses of WATs with respect to content are predominantly qualitative or based on manual evaluation~\cite{ebadi2023investigating}, which introduces the potential for subjective biases~\cite{wibawa2023use}. Prior research tends to examine individual tools, such as Grammarly~\cite{ebadi2023investigating} or Quillbot~\cite{amyatun2023can} rather then providing a comparative analysis, thereby limiting generalizability. With regard to language models, ~\citet{10.1145/3696459} investigate the risk of diversity reduction as a result of language model usage as a case study on the example of ChatGPT.

In contrast to the above works, we focus on a quantitative evaluation of a range of rephrasing tools, with the aim of establishing comparability.

%%%%%%%%%%%%%%%%%%%%%%%%%%%%%%%%%%%%%%%%%%%%%%%%%%%%%%%%
%%%%%%%%%%%%%%%%%%%%%%% Data %%%%%%%%%%%%%%%%%%%%%%%%%%%
%%%%%%%%%%%%%%%%%%%%%%%%%%%%%%%%%%%%%%%%%%%%%%%%%%%%%%%%

\section{Data}

We use a variety of text types that we rephrase with the help of writing assistants and language models, as described in the following.

\subsection{Corpus Compilation}

To compile the corpus, we consider English texts, predominantly written prior to 2010 to exclude those that were created with support by writing assistants or language models. Each individual text has paragraph length, and is selected to be coherent, contiguous, with content relevant to the domain, and not contain non-standard characters that may interfere during rephrasing. To investigate the impact of text type, we compile the corpus from 8 different domains (literature, academic papers, encyclopedic texts, instruction manuals, news, social media posts, interview transcripts, and speeches), resulting in a total of 819 texts.

For further details, see Appendix~\ref{app:textCollection}. The data is available in our code repository\footnote{\href{https://anonymous.4open.science/r/Writing-Assistant-Tools-C661/}{https://anonymous.4open.science/r/Writing-Assistant-Tools-C661/}}.

\subsection{Rephrasing}

Using the corpus as input, we then utilized 4 writing assistants to rephrase these texts. Given recent proliferation of LLMs and their use as writing assistants, we also rephrase via zero-shot prompting with 4 different LLMs.

\noindent
\textbf{Writing assistant tools (WATs).}
As our WATs, we consider the popular tools Grammarly, Quillbot, and Wordtune, and also include Rephrase as a lesser-known tool, all four of which provide rephrasing functionality. For better performance, we obtained membership subscription for all tools. Processing of the input texts was then done manually by one of the coauthors, rephrasing each paragraph independently in the WATs interface. To ensure consistent rephrasing, we adopted the ruleset that (1) all rephrasing suggestions had to be accepted, and (2) in case of multiple rephrasing suggestions, the first (highest ranked) option had to be used. For further details on the tools and rephrasing process, see Appendix~\ref{app:RephraseCriteria}.

\noindent
\textbf{Language model rephrasing (LLMs).}
Since language models are increasingly used for text generation, we also employ four LLMs to rephrase the texts via zero-shot prompting. We consider two closed-source commercial models with ChatGPT (GPT3.5) and GPT4~\cite{achiam2023gpt}, as well as the two pre-trained models Vicuna-13b~\cite{chiang2023vicuna} and Flan-T5 XL~\cite{chung2022scaling}. 

To determine optimal parameter settings and identify the most suitable prompts from a set of prompt candidates, we ran preliminary tests on a small subset of 40 texts that were randomly sampled, and whose results were manually graded. For details on the preliminary experiments, see Appendix~\ref{app:PromptEngineering}, for the used model parameters, see Appendix~\ref{app:parameterSettings}.

%%%%%%%%%%%%%%%%%%%%%%%%%%%%%%%%%%%%%%%%%%%%%%%%%%%%%%%%
%%%%%%%%%%%%%%%%%% Experimental Setup %%%%%%%%%%%%%%%%%%
%%%%%%%%%%%%%%%%%%%%%%%%%%%%%%%%%%%%%%%%%%%%%%%%%%%%%%%%

\section{Experimental Setup}
\label{sec:ExperimentalSetup}

To measure changes to the texts as a result of rephrasing, we consider word-based measures and embedding-based measures, both computed at the paragraph level. For definitions, see Appendix~\ref{app:Metrics}.

\subsection{Word-based Measures}

As word-based metrics, we consider changes at the sentence and the vocabulary level.

\noindent
\textbf{Paragraph length.}  
As the most straightforward measure, we use the percentual change in the length of paragraphs, measured in the number of words. 

\noindent
\textbf{Jaccard similarity.} 
To measure the vocabulary overlap before and after rephrasing, we use the Jaccard similarity~\cite{niwattanakul2013using}.

\subsection{Vector-based Measures}

To measure vector-based changes, we consider the semantic similarity between paragraphs, as well as changes in the size of the cone containing the texts in latent embedding space. % and volume of the cloud of token-level embeddings.

\noindent
\textbf{Semantic similarity.} 
To assess semantic changes incurred during rephrasing, we create paragraph embeddings of texts before and after rephrasing, and compute the cosine similarity.
% semantic similarity as
% \begin{equation}\label{eq:semanticsimilarity}
% sesim(A,B) = cosine(se(A),se(B))
% \end{equation}
We use two models to create embeddings: sentence-BERT~\cite{reimers2019sentence} and ALBERT~\cite{lan2019albert}, with default hyperparameters.

\noindent
\textbf{Conicity.} 
As a metric for assessing the dispersion of a set of vectors, conicity can be applied to measure the spread of token vectors in the latent space of a language model~\cite{sharma2018towards}.
% Fundamentally, it is based on the concept of alignment to the mean (ATM), which is defined for a vector $v \in V$ in a set of vectors as
% \begin{equation}\label{eq:ATM}
% ATM(v,V) = cosine(v,\frac{1}{\left | V \right |}\sum_{x\in V}^{}{x})
% \end{equation}
% The conicity is then defined as the average ATM over all vectors in $V$.
% \begin{equation}\label{eq:conicity}
% conicity(V) = \frac{1}{\left | V \right |}\sum_{v\in V}^{}{ATM(v,V)}
% \end{equation}
Intuitively, if one were to construct the smallest cone that contains all embedding vectors and has its apex at the origin, a larger conicity for a set of vectors denotes a lower spread. A larger conicity value is thus correlated with lower semantic variation in the text.
We obtain token-level embeddings from two different language models to control for possible model bias, namely BERT-Large uncased~\cite{devlin2018bert}, and GPT-2 XL~\cite{radford2019language}. We use the default hyperparameters and extract embeddings from the final hidden layer.

%%%%%%%%%%%%%%%%%%%%%%%%%%%%%%%%%%%%%%%%%%%%%%%%%%%%%%%%
%%%%%%%%%%%%%%%%%%%%%%%% Results %%%%%%%%%%%%%%%%%%%%%%%
%%%%%%%%%%%%%%%%%%%%%%%%%%%%%%%%%%%%%%%%%%%%%%%%%%%%%%%%

\section{Results}

In the following, we show text changes due to rephrasing by writing assistants (WATs) and language models (LLMs) from the perspective of word-based measures and vector-based measures.

\begin{figure}[t]
    \centering
    \includegraphics [width=0.48\textwidth]{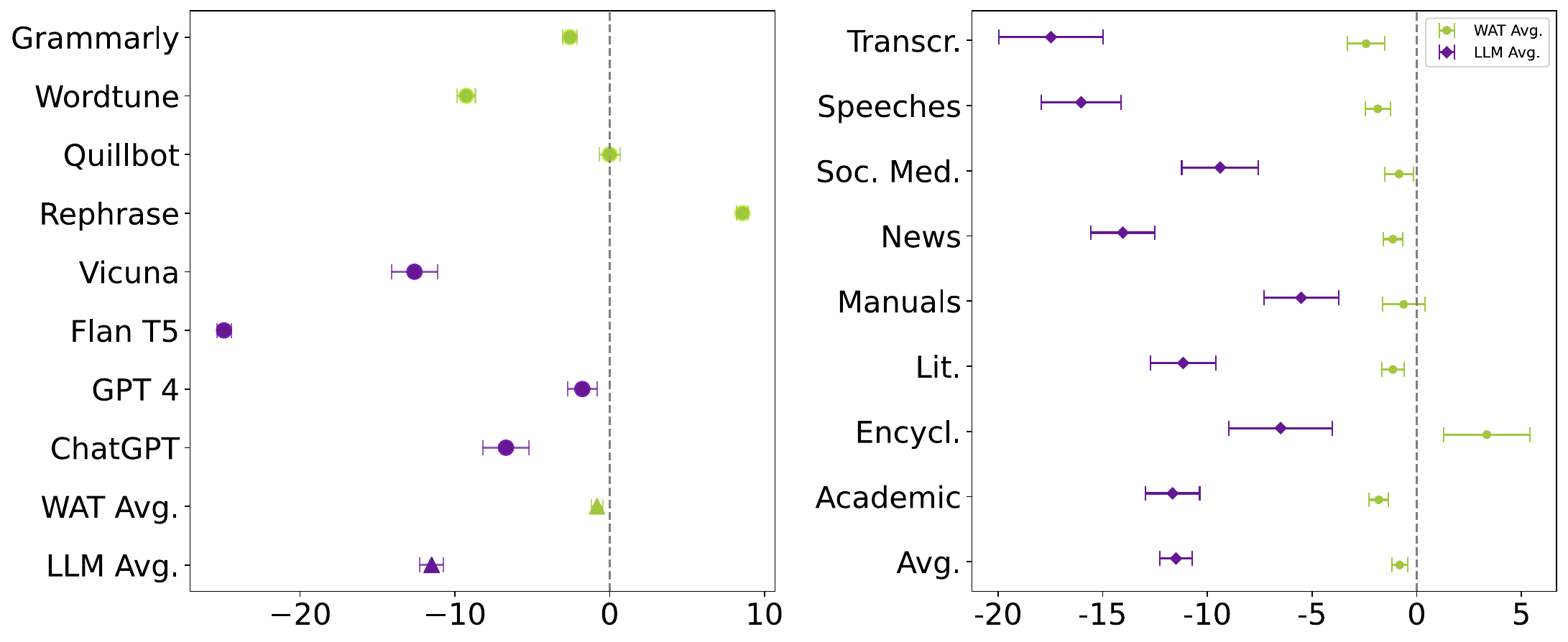}
    \caption{Percentual difference in text length after rephrasing with writing assistants (green) and LLMs (purple). Error bars denote 99\% confidence intervals.}
    \label{fig:Length}
\end{figure}

\subsection{Word-based Measures}

When considering the variation in paragraph length due to rephrasing (see Figure~\ref{fig:Length}, left), we find that, on average, WATs tend to not change the length of texts: while Wordtune shortens the texts by $9.29\%$, rephrase extends them by $8.58\%$, and the other tools enact little change. In contrast, LLMs consistently tend to shorten the texts: FLAN-T5 shortens most drastically by $24.91\%$, while GPT-based models shorten to a lesser degree. Overall, rephrasing by LLMs is much less consistent with respect to length, as their output length varies much more strongly. These findings are consistent across different text types (see Figure~\ref{fig:Length}, right), where we find that all tools tend to shorten texts regardless of type, with the sole exception of encyclopedia texts, which are extended by WATs (in particular: rephrase). % and make up almost all cases of text extension. 
Here as well, WATs shorten the texts to a significantly lesser degree than LLMs.

\begin{figure}[t]
    \centering
    \includegraphics [width=0.48\textwidth]{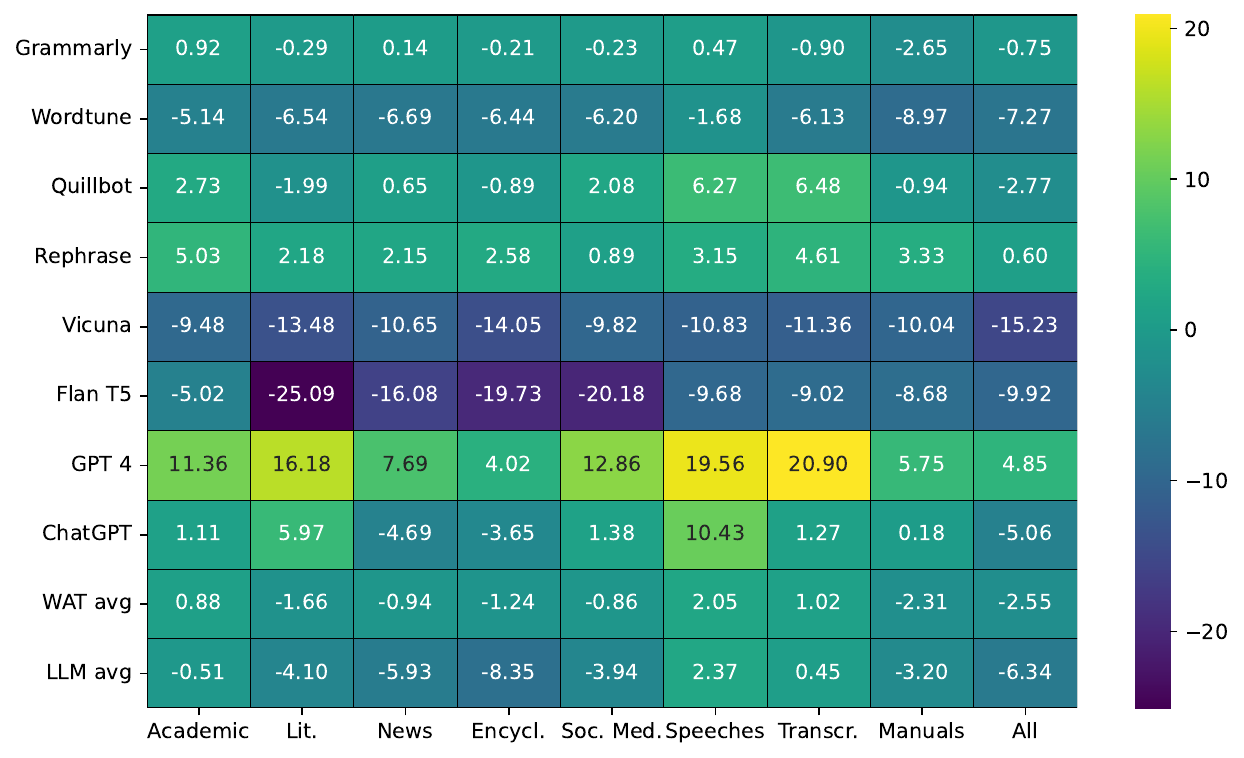}
    \caption{Percentual changes in total vocabulary size between all input and rephrased texts of a given type.} % for all combinations of tool and text type.}
    \label{fig:VocabularySize}
\end{figure}

Considering the vocabulary size (see Figure~\ref{fig:VocabularySize}), we find that WATs consistently decrease the vocabulary size, with the exception of Rephrase. %, which slightly increases it. 
With the exception of GPT-4, LLMs also consistently decrease the vocabulary size, yet more strongly than WATs. The behavior of tools is relatively consistent across text types, although there are variations based on the type. Particularly interesting is the fact that ChatGPT and GPT-4 increase the vocabulary size within domains much more than they do on the corpus level (ChatGPT even decreases it), indicating a strong contraction of linguistic diversity across domains and a change towards the LLMs' own inherent vocabulary.

\begin{figure}[t]
    \centering
    \includegraphics [width=0.48\textwidth]{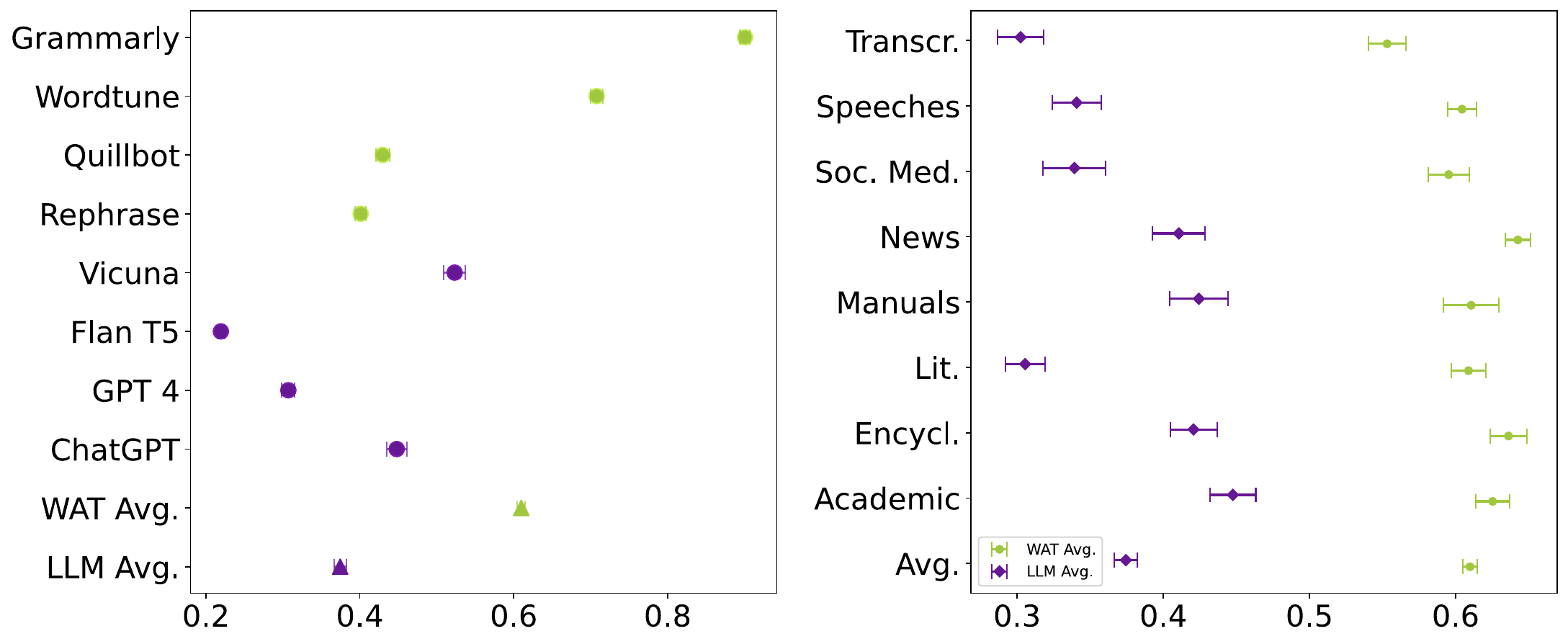}
    \caption{Jaccard overlap of input text and rephrased texts for assistants (green) and LLMs (purple). Error bars denote 99\% confidence intervals.}
    \label{fig:Jaccard}
\end{figure}

With regard to vocabulary overlap (see Figure~\ref{fig:Jaccard}, left), we find the changes resulting from almost all tools to be significant. Only Grammarly (Jaccard score of $0.90$) and Wordtune (Jaccard score of $0.71$) retain a strong overlap with the original texts, while all other tools have Jaccard scores below $0.6$. On average, the change in vocabulary is much stronger for LLMs than it is for writing assistants. 
We again find these results to be consistent across the different text types (see Figure~\ref{fig:Jaccard}, right). %, without exception.

\begin{table}[t]
\small
\centering
\begin{tabular}{llrr}
\toprule
Type & Tool & SBERT & ALBERT \\
\midrule
\multirow{5}{*}{WATs} & Grammarly & 0.9873 & 0.9950 \\
& Wordtune & 0.9539 & 0.9924 \\
& Quillbot & 0.9382 & 0.9883 \\
& Rephrase & 0.8921 & 0.9641\\
& WAT avg. & 0.9429 & 0.9850 \\
\midrule
\multirow{5}{*}{LLMs} & Vicuna & 0.9279 & 0.9824 \\
& FLAN T5 & 0.7484  & 0.9650 \\
& GPT 4 & 0.9240 & 0.9775 \\
& Chat GPT & 0.9260  & 0.9795 \\
& LLM avg. & 0.8816  & 0.9761 \\
\bottomrule
\end{tabular}
\caption{Cosine similarity between paragraph embeddings of original and rephrased texts.}
\label{tab:SemanticChangeAll}
\end{table}

\subsection{Vector-based Measures}

When considering semantic similarity between original and rephrased texts on the basis of vector embeddings (see Table~\ref{tab:SemanticChangeAll}), we find a consistently stronger deviation from the original text for LLMs than for WATs, with the exception of Rephrase. For LLMs, semantic similarity can generally be considered high, the the exception of FLAN-T5, which diverges strongly when we use SBERT embeddings. However, we also find that these differences are potentially model-specific, as ALBERT embeddings consistently indicate very minor semantic changes. % across all models.

These results hold when considering types of text (see Table~\ref{tab:SemanticChangeByType} in Appendix~\ref{app:addResults}), where we observe no outliers for WATs with a much stronger semantic diversion than other types. %, although minor type-dependent differences are visible: speech-based types are edited more strongly, as are social media posts. %Suprisingly, literature also has a relatively higher rate of change by by WATs.
In contrast, rephrasing with LLMs leads to strong divergence overall, in particular for literary text, social media posts, interview transcripts, and speeches. % than it does for news, manuals and academic or encyclopedia texts.

\begin{figure}[t]
    \centering
    \includegraphics [width=0.48\textwidth]{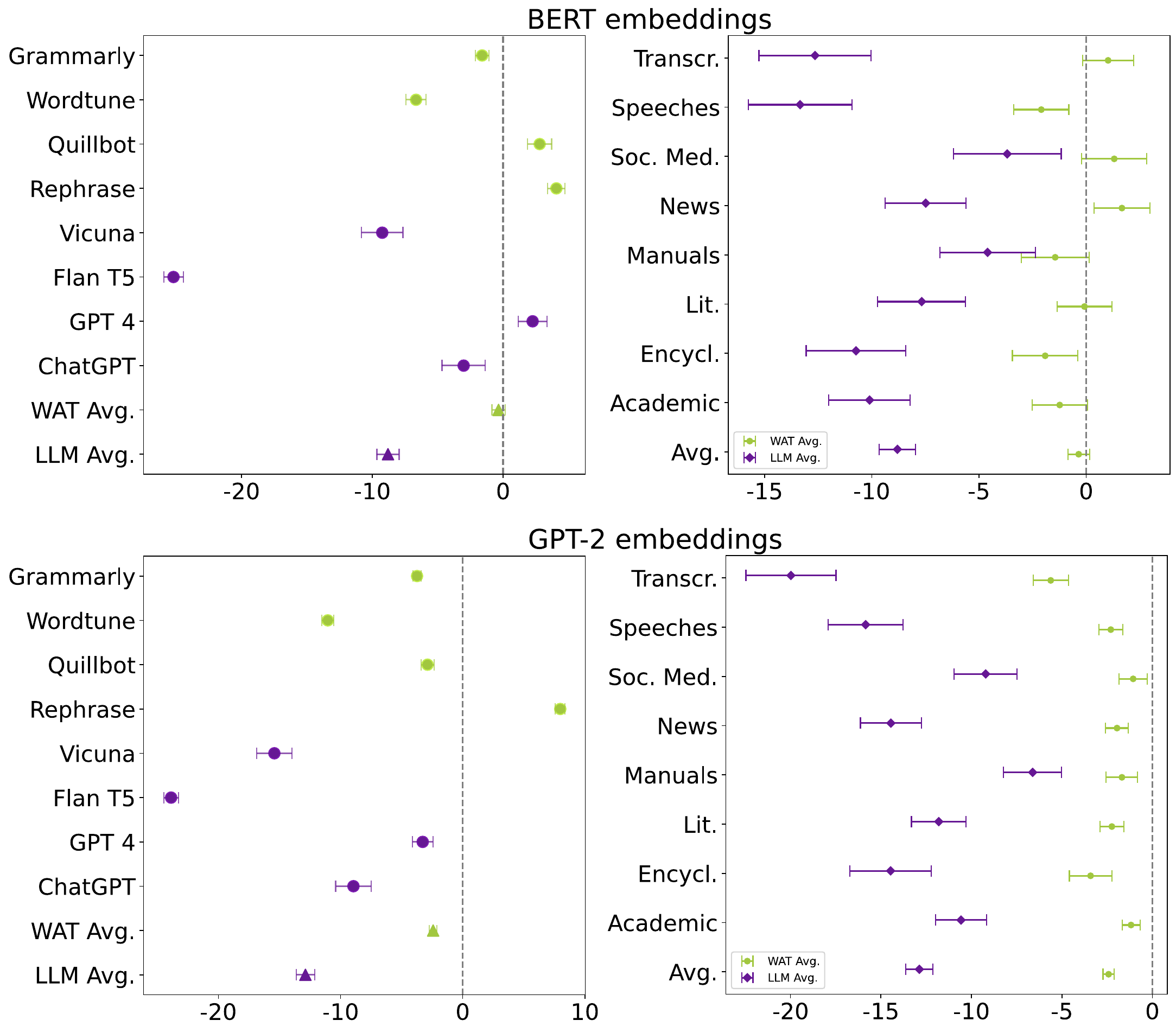}
    \caption{Percentual changes in conicity after rephrasing for WATs (green) and LLMs (purple). Error bars denote 99\% confidence intervals.}
    \label{fig:Conicity}
\end{figure}

To measure the dispersion of rephrased texts in the embedding space, we consider the changes in conicity (see Figure~\ref{fig:Conicity}, left), which indicates that, on average, LLMs cause a narrowing of the semantic spread in the texts, while WATs cause close to no change. This difference is primarily due to the smaller LLMs Vicuna and FLAN-T5, while the GPT-based models' behavior is closer to that of writing assistants. These observations are consistent for both employed embedding models, albeit less pronounced for GPT-2 than for BERT. Overall, the variation in conicity scores is much higher when using LLMs for rephrasing than WATs.
The results are also consistent when considering the type of text (see Figure~\ref{fig:Conicity}, right), where BERT-based embeddings cluster around zero for WATs while they are slightly negative when using GPT-embeddings. In contrast, conicity consistently decreases for LLMs for both embeddings. An interesting difference occurs for transcripts, which show a strong decrease in conicity for LLMs with both embedding models, yet an increase for WATs when using BERT and a decrease using GPT. 

As an alternative to conicity, we also considered changes in the volume of the convex hull as a measure of semantic dispersion, which offered little additional insights (see Appendix~\ref{app:addResults}).

%%%%%%%%%%%%%%%%%%%%%%%%%%%%%%%%%%%%%%%%%%%%%%%%%%%%%%%%
%%%%%%%%%%%%%%%%%% Discussion %%%%%%%%%%%%%%%%%%%%
%%%%%%%%%%%%%%%%%%%%%%%%%%%%%%%%%%%%%%%%%%%%%%%%%%%%%%%%

\section{Discussion and Conclusion}

Our findings indicate three major take-aways.

\subsection{LLMs Are No Writing Assistants}

Based on the semantic similarity between input and rephrased texts, LLMs should not be considered a drop-in replacement for writing assistants. In our experiments, LLMs consistently enact a significantly stronger change in semantics and vocabulary than writing assistants, providing reason to caution against their indiscriminate use, in particular without considering the domain. However, future work should investigate whether this impact can be mitigated through style-sensitive prompting strategies.

\subsection{LLMs Tend to Summarize}

In the comparison between WATs and LLMs, we find that the latter are more strongly inclined to reduce the length of the text, despite the token generation constraint being set well above the length of the input texts. We conjecture that this may be the result of our use of neutral rephrasing prompts in combination with summarization being a likely inclusion during instruction tuning of the models.

\subsection{Reduction of Linguistic Diversity}

Our most drastic finding is the reduction in vocabulary, which is slight but significant for WATs, yet far more pronounced for LLMs. In particular the distribution of strong intra-domain changes in combination with lesser changes at the corpus level suggests that the vocabulary is actively shifted towards the models' internal default vocabulary. This raises concerns of a vocabulary shrinkage and resulting loss in linguistic diversity as a result of WAT and LLM use in text composition. However, similar to ~\citet{10.1145/3696459}, we find that this change is dependent on the model and thereby appears avoidable if suitable design decisions or usage patterns are encouraged, indicating that further research is necessary to prevent an incidental yet avoidable loss of linguistic diversity.

%%%%%%%%%%%%%%%%%%%%%%%%%%%%%%%%%%%%%%%%%%%%%%%%%%%%%%%%
%%%%%%%%%%%%%%%%%% Acknowledgements %%%%%%%%%%%%%%%%%%%%
%%%%%%%%%%%%%%%%%%%%%%%%%%%%%%%%%%%%%%%%%%%%%%%%%%%%%%%%

\section{Limitations}

As a first step in quantifying the impact of using writing assistants and LLMs for rephrasing, our experiments reveal some limitations that should be addressed in future work.

\subsection{Prompting of LLMs and Style}
While we experimented with prompting variations for zero-shot rephrasing with LLMs, we exclusively focused on plain prompts and avoided text style requests such as simplification or domain-specific adaptation. Intuitively, one would expect differences in vocabulary changes when more specific prompts are used, for better or worse.

\subsection{Corpus Limitations}
Although we included a wide range of domains in our corpus, it is far from comprehensive. Future work should expand upon this selection of domains and further investigate domain dependence of linguistic diversity reduction. Similarly, our results are restricted to English, and further languages should be considered.

\subsection{Experimental Scale}
Due to our focus on writing assistants that required semi-manual rephrasing of texts, the size of our corpus is limited to what could feasibly be processed, as it already required days of rephrasing. However, our findings indicate that a larger corpus may be necessary to fully quantify the impact of vocabulary shift as a result of using LLMs. Since there are no constraints to the size of the data for LLM-based rephrasing beyond available compute, future work should investigate this phenomenon with a focus on LLMs on a much larger corpus.

%%%%%%%%%%%%%%%%%%%%%%%%%%%%%%%%%%%%%%%%%%%%%%%%%%%%%%%%
%%%%%%%%%%%%%%%%%% Acknowledgements %%%%%%%%%%%%%%%%%%%%
%%%%%%%%%%%%%%%%%%%%%%%%%%%%%%%%%%%%%%%%%%%%%%%%%%%%%%%%

%\section*{Acknowledgements}

%%%%%%%%%%%%%%%%%%%%%%%%%%%%%%%%%%%%%%%%%%%%%%%%%%%%%%%%
%%%%%%%%%%%%%%%%%% AI Statement %%%%%%%%%%%%%%%%%%%%
%%%%%%%%%%%%%%%%%%%%%%%%%%%%%%%%%%%%%%%%%%%%%%%%%%%%%%%%

\section*{AI Statement}

Language model-based AI tools (ChatGPT) were used as coding assistants in the implementation and as writing assistants in creating a draft of the manuscript. The final version of the manuscript was re-written without AI input.

%%%%%%%%%%%%%%%%%%%%%%%%%%%%%%%%%%%%%%%%%%%%%%%%%%%%%%%%
%%%%%%%%%%%%%%%%%% Bibliography %%%%%%%%%%%%%%%%%%%%
%%%%%%%%%%%%%%%%%%%%%%%%%%%%%%%%%%%%%%%%%%%%%%%%%%%%%%%%

\bibliography{custom}

%%%%%%%%%%%%%%%%%%%%%%%%%%%%%%%%%%%%%%%%%%%%%%%%%%%%%%%%
%%%%%%%%%%%%%%%%%% Appendices %%%%%%%%%%%%%%%%%%%%
%%%%%%%%%%%%%%%%%%%%%%%%%%%%%%%%%%%%%%%%%%%%%%%%%%%%%%%%

\appendix

%%%%%%%%%%%%%%%%%%%%%%%%%%%%%%%%%%%%%%%%%%%%%%%%%%%%%%%%
%%%%%%%%%%%%%%%%%% Data / Text Collection %%%%%%%%%%%%%%
%%%%%%%%%%%%%%%%%%%%%%%%%%%%%%%%%%%%%%%%%%%%%%%%%%%%%%%%

\section{Text Collection}
\label{app:textCollection}

\subsection{Data Sources}

\textbf{Literature.} We randomly select 5 paragraphs from 20 novels written in the 1950s, which are divided between science fiction and romance, and cover authors from the U.S.\ and U.K.\ equally.

\noindent
\textbf{Academic papers.} For the years 2000, 2010, and 2020, we randomly select 7 papers published in NLP from the ACL Anthology\footnote{\href{https://aclanthology.org/}{https://aclanthology.org}} and chose 5 paragraphs from each paper at random.

\noindent
\textbf{Encyclopedia} texts are extracted randomly from articles in Wikipedia\footnote{\href{https://www.wikipedia.org/}{https://www.wikipedia.org}} and the Encyclopedia Britannica\footnote{\href{https://www.britannica.com/}{https://www.britannica.com}} in equal amounts, using 17 keywords each from politics, climate change, and technology that are randomly generated by ChatGPT.

\noindent
\textbf{Instruction Manual} texts cover 20 instruction manuals for electronic and non-electronic products, with 5 paragraphs per manual that we downloaded from Manualsrepo\footnote{\href{https://manualsrepo.com/}{https://manualsrepo.com}}.

\noindent
\textbf{News.} To include news articles, we consider the topics of politics, climate change, and technology. For each, we randomly select 7 articles released between 2011 and 2013 by CNN\footnote{\href{https://edition.cnn.com/sitemap.html}{https://edition.cnn.com/sitemap.html}} and BBC\footnote{\href{https://www.bbc.com/news}{https://www.bbc.com/news}} and extract 5 paragraphs per article.

\noindent
\textbf{Social Media} texts also cover politics, climate change, and technology. For each topic, we chose 17 posts from Instagram and Reddit, using the same randomly generated search terms as for encyclopedias.

\noindent
\textbf{Speeches.} To cover politics, we consider transcripts of State of the Union addresses for 7 U.S.\ presidents. For the topics climate and technology, we use transcripts of 7 TED talks\footnote{\href{https://www.ted.com/}{https://www.ted.com}} each. Per transcript, we extract 5 paragraphs at random.

\noindent
\textbf{Interview Transcripts} were collected from celebrity interviews on Collider\footnote{\href{https://collider.com/}{https://collider.com}}, with half the interview transcripts stemming from native English speakers, and half from non-native speakers.

\subsection{Selection Criteria}

To minimize bias in the data, we defined a set a global criteria applying to all domains, as well as some domain-specific selection and randomization criteria. Global criteria include:
\begin{itemize}
    \item Ensuring that the length of the paragraphs is roughly similar.
    \item Ensuring that no special characters are contained in the texts.
    \item Ensuring that no names appear in the text.
    \item Ensuring that the text is contiguous and relevant to the domain.
    \item Restricting the text to English content.
\end{itemize}

As domain-specific selection criteria, we also consider the following.

\noindent
\textbf{Literature.} Paragraphs are sampled randomly from the entire book, such that each text has roughly the same length and contains ten sentences. If necessary, adjacent paragraphs are merged to create a text of sufficient length. The inclusion of fictional names is avoided.

\noindent
\textbf{NLP Papers.} We exclude the abstract of papers from the selection. We avoid the inclusion of formulae or mathematical characters. Texts are selected to be roughly ten sentences long and from contiguous sequences of text in the paper.

\noindent
\textbf{Encyclopedias.} We avoid text with non-standard symbols or characters. All text samples possess similar lengths and numbers of sentences.

\noindent
\textbf{Instruction Manuals.} Several neighboring paragraphs from the same section are selected to create each text sample. The number of sentences and the length are kept comparable.We avoid sections that strongly rely on numbered instructions indicating steps or sequences.

\noindent
\textbf{News.} Texts are chosen to contain content relevant to the selected topic. Each text is a contiguous segment from the article with a consistent length and amount of sentences.

\noindent
\textbf{Social Media.} Each paragraph is taken from a complete comment or discussion. Paragraphs including non-English phrases are avoided. Comparable text lengths and identical number of sentences is ensured, with no non-standard symbols.

\noindent
\textbf{Speeches.} Texts are chosen to represent a consistent response to the same issue. Annotations such as applause and laughing from the audience are removed manually. We ensure similar text length and number of sentences, with no non-standard symbols.

\noindent
\textbf{Interview Transcripts.} We select coherent responses or a question with a subsequent response to constitute each text.

\subsection{Data Composition}

The final corpus consists of 819 paragraph-length texts. For an overview, see Table~\ref{Dataset Structure table}.

\begin{table}
    \centering
    \small
    \begin{tabular}{lrrr}
        \toprule
Type & Politics&  Climate& Technology\\
        \midrule
News&  35&  35& 35\\
Encycl. &  34&  34& 34\\
Soc. Med.&  34&  34& 34\\
Speeches&  35&  35& 35\\
         \toprule
         &  2020&  2010& 2000\\
         \midrule
Academic &  35&  35& 35\\
         \toprule
          & Romance & SciFi & \\
          \midrule
 Lit. & 50 & 50 & \\
         \toprule
         &  Entertainment &  Tech &\\
         \midrule
Transcr. &  50 & 50 & \\
         \toprule
         &  Electr. &  Mech. &\\
         \midrule
Manuals&  50&  50 &    \\
         \bottomrule
    \end{tabular}
    \caption{Counts of texts by type. Each type contributes roughly 100 texts to the corpus.}
    \label{Dataset Structure table}
\end{table}

%%%%%%%%%%%%%%%%%%%%%%%%%%%%%%%%%%%%%%%%%%%%%%%%%%%%%%%%
%%%%%%%%%%%%%%%%%% Rephrasing %%%%%%%%%%%%%%%%%%%%
%%%%%%%%%%%%%%%%%%%%%%%%%%%%%%%%%%%%%%%%%%%%%%%%%%%%%%%%

\section{Rephrasing Criteria}
\label{app:RephraseCriteria}

For rephrasing texts with writing assistants, we follow a set of general criteria. All writing assistants are used in their default settings. Each paragraph is rephrased separately to maintain the original meaning and structure, without reformatting citations or lists. The length of paragraphs is designed to allow us rephrasing without having to split texts.

Additionally, we use specific settings for each of the assistants to make their performance as close as possible, despite differing features. We also list our settings for the UI-based interaction with ChatGPT here.
\begin{itemize}
    \item \textbf{Grammarly\footnote{\href{https://app.grammarly.com/}{https://app.grammarly.com/}}} All services are utilized. Rephrasing done sequentially from the start to the end of the text. The first option is always chosen, and no reformatting of citations or lists performed. Paragraphs are not split, and rephrasing continues until no further suggestions are made. Extra attention is given to speech and interview texts to avoid the inclusion of recurring, alternating patterns that is sometimes suggested by Grammarly.

    \item \textbf{Wordtune\footnote{\href{https://www.wordtune.com/rewrite}{https://www.wordtune.com/rewrite}}} All services are used, and rephrasing is done by paragraph, not sentence by sentence. The first suggestion is always selected, and rephrasing continues until no further suggestions are provided. Original content is kept for sentences that are too long for the tool to to offer advice.

    \item \textbf{Quillbot\footnote{\href{https://quillbot.com}{https://quillbot.com}}} All services are used, and the first generated content is accepted without any manual modifications.

    \item \textbf{Rephrase\footnote{\href{https://www.rephrase.info/}{https://www.rephrase.info}}} All services are used, and the first generated content is accepted without any manual modifications.

    \item \textbf{ChatGPT.} The basic service is used, ensuring paragraph structure is preserved. The prompt we use is \texttt{"Could you help me to rephrase this paragraph: $\langle$input$\rangle$"}. Warnings about sensitive information (e.g., politics) are ignored, and when multiple responses are provided, the first option is always selected. In cases where the model provides a response that is identical to the input, we accept it.
\end{itemize}

%%%%%%%%%%%%%%%%%%%%%%%%%%%%%%%%%%%%%%%%%%%%%%%%%%%%%%%%
%%%%%%%%%%%%%%%%%% Prompt Engineering %%%%%%%%%%%%%%%%%%%%
%%%%%%%%%%%%%%%%%%%%%%%%%%%%%%%%%%%%%%%%%%%%%%%%%%%%%%%%

\section{Prompt Engineering}
\label{app:PromptEngineering}

Since the outputs of LLMs varies based on the used prompt template, we experimented with multiple possible prompts for rephrasing, shown in Table~\ref{tab:prompts}. 

\begin{table}
%\centering
\small  % Reduce font size for the table
\begin{tabular}{llp{5cm}}  % Adjust column width to fit half the page width
\toprule 
Model & ID & Prompt \\
\midrule
\multirow{6}{*}{Flan-T5} 
& 1 & In need of some rephrasing expertise. Can you help me rephrase this paragraph? $\langle$input$\rangle$ \\
& 2 & Could you help me to rephrase this paragraph with similar length: $\langle$input$\rangle$ \\
& 3 & user: rephrase the original text: $\langle$input$\rangle$ assistant: \\
& 4 & I'm not quite satisfied with how this paragraph sounds. Can you rephrase it? $\langle$input$\rangle$ \\
& 5 & Please rephrase the following paragraph: $\langle$input$\rangle$ Rephrased paragraph: \\
& 6 & Please rephrase the following paragraph. Preserve the meaning but change the words: $\langle$input$\rangle$ Rephrased paragraph: \\
\midrule
\multirow{6}{*}{Vicuna}
& 1 & user: rephrase the original text: $\langle$input$\rangle$ assistant: \\
& 2 & Can you assist me in rephrasing the following paragraph: $\langle$input$\rangle$ \\
& 3 & Could you provide a rephrased version of the following paragraph: $\langle$input$\rangle$ \\
& 4 & How would you rephrase the following paragraph: $\langle$input$\rangle$ \\
& 5 & Please rephrase the following paragraph: $\langle$input$\rangle$ Rephrased paragraph: \\
& 6 & Please rephrase the following paragraph. Preserve the meaning but change the words: $\langle$input$\rangle$ + "Rephrased paragraph:" \\ 
\bottomrule
\end{tabular}
\caption{Prompt candidates for Flan-T5 and Vicuna.}
\label{tab:prompts}
\end{table}

\subsection{Prompt Selection}

Using a small test set of 40 randomly selected paragraphs from our corpus, we prompted both models with all six prompts and manually graded the output with a score from 0 to 4 using the following scheme:
\begin{itemize}
\item[0] If the model made no change to the text or produced empty output.
\item[1] If the model summarizes the input to 1-2 sentences or begins with "This paragraph described".
\item[2] If the model shortens the text, removes the last few sentences, or keeps sentences without revision.
\item[3] If the model rephrases but adds or deletes sentences.
\item[4] If the model rephrases and retains the original length.
\end{itemize}
For each model, we selected the prompt with the highest total score across all outputs.

\subsection{Experiment Prompts}

Based on this preliminary experiment, the prompts that we use in the experiments are \texttt{"Could you help me to rephrase this paragraph:$\langle$input$\rangle$"} for ChatGPT and \texttt{"Please rephrase the following paragraph. Preserve the meaning but change the words:$\langle$input$\rangle$.\ Rephrased paragraph:"} for Vicuna, Flan-T5, and GPT-4.

%%%%%%%%%%%%%%%%%%%%%%%%%%%%%%%%%%%%%%%%%%%%%%%%%%%%%%%%
%%%%%%%%%%%%%%%%%% Experiment Parameters %%%%%%%%%%%%%%%
%%%%%%%%%%%%%%%%%%%%%%%%%%%%%%%%%%%%%%%%%%%%%%%%%%%%%%%%

\section{Experiment Parameter Settings}
\label{app:parameterSettings}

Using the test sample we employed for the selection of prompts, we also tested a number of parameter settings to determine which hyperparameters had the highest change of providing suitable rephrased texts. Based on these tests, we use as hyperparameter settings:

\noindent
\textbf{Vicuna.}
We used version vicuna-13b-v1.5. The model was configured with \texttt{num\_return\_sequences = 1}, \texttt{max\_length = 1024}, and disabled sampling. This setup ensured the generation of a single sequence constrained to a maximum length of 1024 tokens.

\noindent
\textbf{Flan-T5.}
We used the version flan-t5-xl. The model was configured with \texttt{min\_length} set to 90\% of the original length, \texttt{max\_length} set to 110\% of the original length, temperature set to 1.6, sampling enabled, and \texttt{top\_k = 25}.

\noindent
\textbf{GPT4.}
For the GPT-4 model we set the temperature parameter to 1.

%%%%%%%%%%%%%%%%%%%%%%%%%%%%%%%%%%%%%%%%%%%%%%%%%%%%%%%%
%%%%%%%%%%%%%%%%%% Metric Details %%%%%%%%%%%%%%%%%%%%
%%%%%%%%%%%%%%%%%%%%%%%%%%%%%%%%%%%%%%%%%%%%%%%%%%%%%%%%

\section{Metric Definitions}
\label{app:Metrics}

In the following, we provide further details on the metrics used in the experiments.

\subsection{Paragraph length} 
For texts before rephrasing ($B$) and after rephrasing ($A$), we compute
\begin{equation}\label{eq:length}
length(A,B) = \frac{tokens(A) - tokens(B)}{tokens(B)}
\end{equation}
with negative values denoting that the number of tokens decreased due to rephrasing. For tokenization, we use NLTK.

\subsection{Jaccard similarity}
To measure the vocabulary overlap between paragraphs before and after rephrasing, the Jaccard similarity treats both texts as sets of words and computes the overlap as:
\begin{equation}\label{eq:jaccard}
Jaccard(A,B) = \frac{|A \cap B|}{|A \cup B|}
\end{equation}
with a value of $1$ denoting a perfect match in the vocabulary.
For preprocessing, we use NLTK to tokenize, convert letters to lowercase, remove punctuation marks, and eliminate stopwords.

\subsection{Semantic similarity}
To assess the semantic changes incurred during rephrasing, we create embeddings $se$ of paragraphs before ($B$) and after ($A$) rephrasing, and compute the semantic similarity as
\begin{equation}\label{eq:semanticsimilarity}
sesim(A,B) = cosine(se(A),se(B))
\end{equation}

\subsection{Conicity}
We use the definition of conicity by~\citet{sharma2018towards}. It is based on the concept of alignment to the mean (ATM), which is defined for a vector $v \in V$ in a set of vectors as
\begin{equation}\label{eq:ATM}
ATM(v,V) = cosine(v,\frac{1}{\left | V \right |}\sum_{x\in V}^{}{x})
\end{equation}
The conicity is then defined as the average ATM over all vectors in $V$.
\begin{equation}\label{eq:conicity}
conicity(V) = \frac{1}{\left | V \right |}\sum_{v\in V}^{}{ATM(v,V)}
\end{equation}

\subsection{Convex Hull Volume}

As an alternative to conicity, we also consider the volume of the convex hull (ch), which has been used to assess assessing the semantic content of textual data~\cite{yu2022can, ma2023conceptual}. Formally, the convex hull of a finite set of points in Euclidean space is defined as the minimal polygon encompassing all points. For texts before rephrasing ($B$) and after rephrasing ($A$), we compute the percentual change in the volume of the convex hull as
\begin{equation}\label{eq:convexhull}
cch(A,B) = \frac{vol(ch(A)) - vol(ch(B))}{vol(ch(B))}
\end{equation}
Intuitively, an increase in the volume of the convex hull indicates that the rephrased text has stronger semantic variation than the original.

To avoid degenerate results due to varying and low dimensions (the number of words in the considered texts is generally lower than the embedding dimension), we use PCA to reduce the dimension of embeddings to $8$. For computing the convex hulls, we utilize the Quickhull Algorithm~\cite{barber1996quickhull}.

%%%%%%%%%%%%%%%%%%%%%%%%%%%%%%%%%%%%%%%%%%%%%%%%%%%%%%%%
%%%%%%%%%%%%%%%%%% Additional Results %%%%%%%%%%%%%%%%%%%%
%%%%%%%%%%%%%%%%%%%%%%%%%%%%%%%%%%%%%%%%%%%%%%%%%%%%%%%%

\section{Additional Results}
\label{app:addResults}

\subsection{Semantic Similarity by Type}
\label{app:SemSimType}

In Table~\ref{tab:SemanticChangeByType}, we show cosine similarities between paragraph embeddings before and after rephrasing, grouped by the type of text.

\begin{table*}[t]
\small
\centering
\begin{tabular}{llrrrrrrrr}
\toprule
Type & Tool & Academic & Lit. & News & Encycl. & Soc. Med. & Speeches & Transcr. & Manuals \\
\midrule
\multirow{10}{*}{SBERT} 
& Grammarly & 0.9866 & 0.9863 & 0.9943 & 0.9935 & 0.9871 & 0.9857 & 0.9794 & 0.9855\\
& Wordtune & 0.9783 & 0.9497 & 0.9572 & 0.9647 & 0.9436 & 0.9392 & 0.9368 & 0.9610\\
& Quillbot & 0.9672 & 0.9116 & 0.9518 & 0.9532 & 0.9314 & 0.9217 & 0.9185 & 0.9485\\
& Rephrase & 0.8563 & 0.9091 & 0.9252 & 0.9047 & 0.8963 & 0.8803 & 0.8624 & 0.9028\\
& WAT avg. & 0.9471 & 0.9392 & 0.9571 & 0.9540 & 0.9396 & 0.9317 & 0.9243 & 0.9495\\
\cmidrule{2-10}
& Vicuna & 0.9711 & 0.9026 & 0.9438 & 0.9615 & 0.9115 & 0.8855 & 0.8930 & 0.9526\\
& Flan T5 & 0.7765 & 0.7307 & 0.7907 & 0.7846 & 0.7274 & 0.7316 & 0.6744 & 0.7682\\
& ChatGPT & 0.9648 & 0.9073 & 0.9391 & 0.9467 & 0.9174 & 0.8982 & 0.8855 & 0.9473\\
& GPT 4. & 0.9445 & 0.9006 & 0.9495 & 0.9571 & 0.9103 & 0.8971 & 0.8922 & 0.9395\\
& LLM avg. & 0.9142 & 0.8603 & 0.9058 & 0.9125 & 0.8667 & 0.8531 & 0.8363 & 0.9019\\
\bottomrule
\end{tabular}
\caption{Cosine similarity between paragraph embeddings of original and rephrased texts, grouped by type.}
\label{tab:SemanticChangeByType}
\end{table*}

\subsection{Convex Hull Results}
\label{app:convexHull}

\begin{figure}[t]
    \centering
    \includegraphics [width=0.48\textwidth]{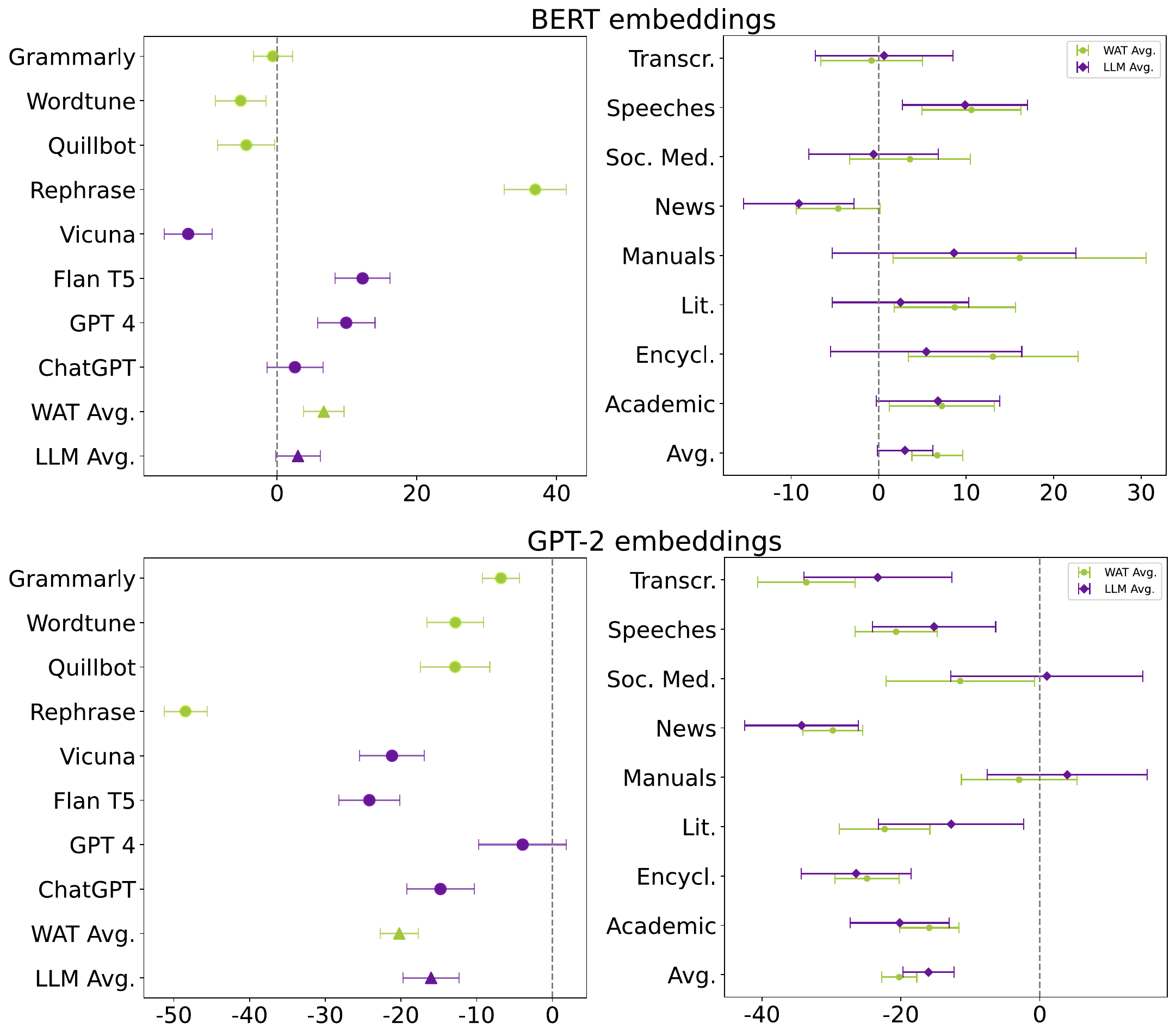}
    \caption{Percentual changes in the volume of the complex hull between input and rephrased texts using BERT and GPT-2 embeddings for assistants (green) and LLMs (purple). Error bars denote 99\% confidence intervals.}
    \label{fig:ConvexHull}
\end{figure}

When considering the convex hull as a measure of semantic dispersion (see Figure~\ref{fig:ConvexHull}) we observe little consistent change regardless of the used tool and embedding model. In almost all cases, the changes are consistent between the rephrasing tools yet vary strongly in the covered area, indicating that conicity is a much more suitable measure of semantic dispersion.

\section{Dataset Structure}
\label{Dataset Structure}

\end{document}